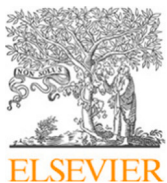
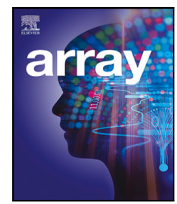

# UDGS-SLAM: UniDepth Assisted Gaussian Splatting for Monocular SLAM

Mostafa Mansour [a],*, Ahmed Abdelsalam [b],*, Ari Happonen [b], Jari Porras [b,c], Esa Rahtu [d]

[a] *Faculty of Engineering and Natural Sciences, Tampere University, Finland*
[b] *School of Engineering Science, LUT University, Finland*
[c] *School of Electrical Engineering, Aalto University, Finland*
[d] *Faculty of Information Technology and Communication Sciences, Tampere University, Finland*

ABSTRACT

Recent advancements in monocular neural depth estimation, particularly those achieved by the UniDepth network, have prompted the investigation of integrating UniDepth within a Gaussian splatting framework for monocular SLAM. This study presents UDGS-SLAM, a novel approach that eliminates the necessity of RGB-D sensors for depth estimation within Gaussian splatting framework. UDGS-SLAM employs statistical filtering to ensure local consistency of the estimated depth and jointly optimizes camera trajectory and Gaussian scene representation parameters. The proposed method achieves high-fidelity rendered images and low ATE-RMSE of the camera trajectory. The performance of UDGS-SLAM is rigorously evaluated using the TUM RGB-D dataset and benchmarked against several baseline methods, demonstrating superior performance across various scenarios. Additionally, an ablation study is conducted to validate design choices and investigate the impact of different network backbone encoders on system performance.

## 1. Introduction

Visual Simultaneous Localization and Mapping (VSLAM) is essential for estimating the pose of a moving vision sensor while simultaneously constructing a map of the environment. VSLAM plays a crucial role in various applications, such as robotics, virtual reality, and augmented reality [1–3]. The choice of map (or scene) representation significantly affects the performance of the SLAM system, influencing both its internal subsystems and external systems that rely on its outputs.

The representation of the map has been a focal point of extensive research [4]. Approaches have focused on explicit handcrafted sparse [5–8] and dense representations [9–12], utilizing points, voxels, surfels, and signed distance fields for map construction. However, despite their maturity, these representations have notable limitations. They heavily depend on the availability of 3D geometric features and are limited to representing only observed parts of the environment. Furthermore, they lack the ability to generate or synthesize photorealistic, high-fidelity novel scenes from different camera viewpoints, which is critical in virtual and augmented reality applications.

To address these limitations, recent research has focused on implicit volumetric photorealistic representations, such as Neural Radiance Fields (NeRF) [13] and Gaussian Splatting (GS) [14], to enable unified, high-fidelity scene representation [15]. NeRF-based methods enhance scene representation by minimizing photometric losses through differentiable rendering, encoding the scene into the weight space of a neural network using multi-layer perceptrons and ray marching [13,15–18]. However, NeRF methods face several challenges, including high computational requirements, long training times, overfitting to specific objects or scenes, and susceptibility to catastrophic forgetting [15]. In contrast, GS uses tile-based rasterization for efficient rendering and represents the scene using 3D Gaussians [14], optimized with each new input. This approach is computationally more efficient, adapts well to large scenes, and avoids catastrophic forgetting. Additionally, GS-based SLAM integrates both photometric information and depth maps, making it better suited for explicit spatial geometry modeling. The differentiable rendering of Gaussians, combined with rapid GPU-based implementation, allows for fast and joint optimization of scene parameters and camera trajectory.

Due to these advantages, GS has emerged as a leading method for photorealistic 3D scene reconstruction, unifying representations for tracking, mapping, and rendering. Recent research predominantly focuses on utilizing RGB-D inputs to benefit from depth sensors [19–25]. However, there remains a gap in exploring monocular methods due to the lack of depth information [19].

Inspired by recent advancements in monocular depth estimation through neural networks [26], we investigate the use of neural depth estimation within the GS framework for monocular SLAM. This approach eliminates the need for RGB-D sensors while retaining the benefits of GS for scene representation. Specifically, we explore the use of the UniDepth network [27] for depth estimation in the GS-based monocular SLAM framework. Our proposed method emphasizes the joint optimization of camera trajectory and 3D Gaussian map representation, utilizing the depth estimation from the UniDepth network. Additionally, we introduce a statistical filtering technique to enhance the local consistency of the estimated depth, improving the quality of photorealistic reconstruction.

In summary, our contributions are as follows:

- **Integrating UniDepth network for depth estimation within Gaussian Splatting**: We leverage UniDepth for depth estimation from RGB images within the Gaussian Splatting framework, facilitating the joint optimization of camera trajectory and 3D photorealistic reconstruction.
- **Introducing Statistical Filtering**: We implement a straightforward yet effective statistical filtering stage that ensures local consistency of the estimated depth map, enhancing the overall performance of the framework.
- **Evaluation on real dataset benchmark**: Our method is rigorously tested on TUM RGB-D dataset, demonstrating superior performance compared to baseline methods in various scenarios.
- **Ablation studies**: We conduct comprehensive ablation studies using different backbone encoders of the UniDepth network, both with and without the implementation of statistical filtering, to evaluate their impact on performance.

## 2. Related work

### 2.1. Monocular neural depth estimation

Monocular depth estimation (MDE) is a classic research area in computer vision that involves determining the precise depth of each pixel in an image. This capability enables the reconstruction of 3D scenes from 2D images, which is crucial for a wide range of applications in computer vision and robotics, such as autonomous navigation, augmented reality, object detection, and 3D modeling. Early methods relied heavily on geometric principles and handcrafted features to estimate depth [28–30].

The advent of deep learning revolutionized the field of computer vision, including MDE. Neural networks offer a data-driven approach to learning complex features directly from images, allowing depth estimation from a single image. One of the earliest neural network-based methods for MDE was introduced by Eigen et al. [31]. This method leveraged the ability of CNNs to capture hierarchical features, achieving reasonable accuracy on the NYU [32] and KITTI [33] datasets. As neural networks continued to evolve, two main branches of monocular depth estimation from a single image emerged: Monocular Metric Depth Estimation (MMDE) and Monocular Relative (Scale-Agnostic) Depth Estimation (MRDE).

Focusing on MMDE [31,34–41], it aims to predict absolute values in physical units (e.g., meters), which is necessary to perform 3D reconstruction effectively. Most of the existing MMDE methods have shown great accuracy across several benchmarks, but they fail to generalize to real-world scenarios and tend to overfit specific datasets [42]. Some methods have attempted to solve this by training a single metric depth estimation model across multiple datasets, but it has been reported that this often deteriorates performance, especially when the collection includes images with large differences in depth scale, such as indoor and outdoor images [42].

Few methods [43,44] have tackled the challenging problem of generalization, but these methods still rely on controlled testing conditions, including fixed camera intrinsics. Hu et al. relaxed some of these conditions; however, their solution still requires known camera intrinsics [45]. Unlike other methods, UniDepth [46] addresses the generalization problem without the limitation of fixed camera intrinsics. UniDepth consistently sets new state-of-the-art benchmarks, even compared with non-zero-shot methods. Therefore, the UniDepth network is chosen for depth estimation as a first step in our pipeline.

### 2.2. NeRF based SLAM

Mildenhall et al. introduced NeRF as an implicit volumetric scene representation [47]. Originally, NeRF required known camera poses to construct its scene representation. To accommodate this, many studies have employed the COLMAP structure-from-motion package [48] to estimate camera poses for use in NeRF implementations. iMAP [49] was the first to relax the requirement for known camera poses by simultaneously performing tracking and mapping using NeRF representation. Despite its innovation, iMAP faced scalability issues, which were subsequently addressed by NICE-SLAM [17] through the introduction of hierarchical multi-feature grids. Vox-Fusion [18] proposed a hybrid solution that combines NeRF with traditional volumetric fusion methods to enhance scene representation. Recently, Point-SLAM [50] enhanced 3D reconstruction by employing neural point clouds and feature interpolation for volumetric rendering. Other additional improvements are discussed in Tosi et al. [15]. Despite these advancements, NeRF-based methods still fundamentally grapple with long training times due to the computational demands of ray marching rendering, and issues with catastrophic forgetting. In contrast, Gaussian-based methods avoid these pitfalls by incorporating dynamic insertion and pruning techniques to manage newly visible scenes, and by utilizing fast rasterization instead of ray marching to boost rendering efficiency.

### 2.3. 3D Gaussians based SLAM

Since its introduction as a promising 3D scene representation [14], 3D Gaussian splatting has emerged as a prominent technology for SLAM due to its fast rasterization rendering via splatting and its ability to overcome the catastrophic forgetting problem through Gaussian insertion and pruning management [15]. Given the importance of depth maps in Gaussian representation, most studies employ RGB-D cameras within the Gaussian framework, benefiting from the integration of depth sensors [15,19,20,25,51–55]. Li et al. focused on generating novel syntheses from sparse inputs by optimizing radiance fields with depth regularization, without optimizing the camera trajectory [56]. Unlike them, UDGS-SLAM optimizes both the camera trajectory and scene representation. Additionally, UDGS-SLAM proposes a statistical filter to enforce local consistency and geometric stability. Some works have integrated RGB-D cameras with IMUs within the Gaussian splatting framework [54], while others have incorporated various depth and normal prior cues with RGB-D measurements [51]. However, there is a notable lack of investigation into using monocular camera measurements within the Gaussian splatting framework, primarily, due to the absence of direct depth measurement.

Lee et al. proposed a monocular-based 3D Gaussian representation for scenes without estimating camera trajectory [57]. They utilized IMU and LiDAR data to provide odometry and estimate the trajectory. In contrast, UDGS-SLAM does not rely on any complementary sensors for trajectory estimation and uses only a monocular camera for this purpose. Matsuki et al. introduced Gaussian splatting for SLAM using a monocular camera [19]. Their approach utilized prior knowledge about scene depth, initializing the 3D Gaussians with depths normally distributed around the mean scene depth. To address the lack of direct depth sensor data, they optimized the Gaussian parameters and camera trajectory by minimizing the photometric error. In contrast, UDGS-SLAM does not rely on any prior knowledge about scene depth. It leverages statistically filtered depth maps from the UniDepth network for initialization. Furthermore, it optimizes the Gaussian parameters and camera trajectory by minimizing a weighted sum of photometric and geometric errors. This approach enables UDGS-SLAM to outperform the monocular SLAM proposed by Matsuki et al. in most scenarios of the TUM dataset, as presented in Section 5.

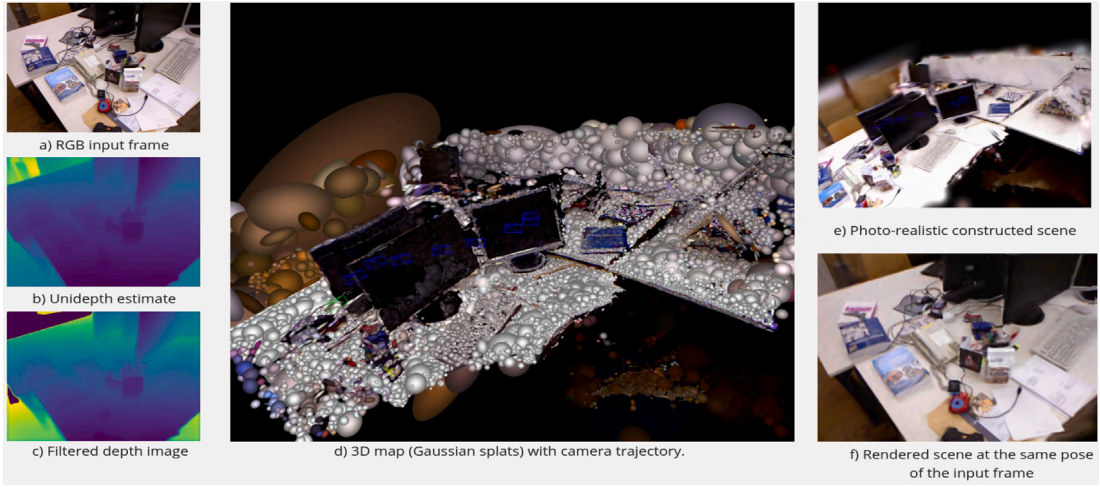

**Fig. 1.** *UDGS-SLAM* utilizes 3D Gaussian splats for scene representation, enabling high-fidelity photorealistic reconstruction for dense SLAM using a monocular camera. It employs the Unidepth network to estimate scene depth from a single RGB image (a, b). The estimated depth is subsequently filtered for local consistency (c). Through differential rendering rasterization, it generates rendered RGB and depth images for a given camera pose. The system then achieves 3D scene representation by jointly optimizing 3D Gaussian splats and camera trajectory through the minimization of photometric error between the input and rendered RGB images, as well as the minimization of geometric error between the estimated and rendered depths (d). This approach enables the reconstruction of a dense scene (e) and allows for photorealistic rendering of the scene from any given camera pose (f).

## 3. Methodology

The proposed approach estimates the camera poses for each frame $\{P_i\}_{i=1}^N$ and reconstruct a 3D volumetric map representation of the scene from a sequential RGB image stream $\{I_i\}_{i=1}^N$ obtained from a monocular camera with known camera intrinsic $\mathbf{K} \in \mathbb{R}^{3 \times 3}$. The map is represented by a collection of 3D Gaussians, which can be rendered into a photorealsitic image for a given view point of a camera pose. This representation is achieved by using differentiable rendering through 3D Gaussian splatting and gradient-based optimization, facilitating the optimization of the camera pose for each frame as well as the volumetric representation of the scene.

### 3.1. 3D Gaussian scene representation

The proposed approach optimizes the scene representation to effectively capture both geometrical and appearance features, enabling it to be rendered into high-fidelity color and depth images. We represent the scene as a set of 3D Gassians coupled with view-independent color, opacity, and a covariance matrix.

$$\mathbf{G} = \{G_i : (\mu_i^W, \mathbf{c}_i, o_i, \Sigma_i^W) \mid i = 1, \ldots, N\}. \quad (1)$$

Each 3D Gaussian $G_i$, in the world coordinate frame $W$ is defined by its center position $\mu_i^W \in \mathbb{R}^3$, its RGB color $\mathbf{c}_i$, a covariance matrix $\Sigma_i^W$, and its opacity $o \in [0, 1]$. A Gaussian $G_i$ affects a 3D point $X \in \mathbb{R}^3$ according to the Gaussian equation weighted by its opacity as follows:

$$f(X) = o_i \left( \frac{\exp(-\frac{1}{2}(X - \mu_i^W)^T \Sigma_i^{W-1}(X - \mu_i^W))}{(2\pi)^{3/2} \left| \Sigma_i^W \right|^{1/2}} \right). \quad (2)$$

### 3.2. Color and depth differentiable rendering via splatting

The objective of Gaussian splatting [20] is to render high-fidelity RGB and depth images from the 3D volumetric Gaussian scene representation given a camera pose. Importantly, the rendering should be differentiable allowing the gradient to be calculated for the underlying Gaussians' map parameters and camera poses with respect to the photometric and geometric discrepancies between the rendered and the provided RGB and depth images, respectively. The gradient is used to minimize the discrepancies by updating both the parameters of 3D Gaussian splats and camera poses. According to [14], an RGB image is rendered from a set of 3D Gaussians by, first, sorting all the Gaussians from front to back with respect to a given camera pose. Then, the 3D Gaussians within the camera frustum are splatted (projected) into 2D pixel space using the camera pose and the camera intrinsic matrix $K$. Finally, an RGB image can be rendered by alpha-blending each 2D splatted Gaussian in order in pixel space. The rendered color of a $p = (u, v)$ pixel can be written as:

$$C(p) = \sum_{i=1}^n \mathbf{c}_i f(p) \prod_{j=1}^{i-1}(1 - f(p)), \quad (3)$$

where $n$ is the number of pixels per image and $f(p)$ is calculated according to (2) after projecting each 3D Gaussian $\mathcal{N}(\mu^W, \Sigma^W)$ into 2D image space Gaussian $\mathcal{N}(\mu^I, \Sigma^I)$ as follows:

$$\mu^I = \pi(T_{CW} \mu^W),$$
$$\Sigma^I = \mathbf{JR}\Sigma^W(\mathbf{JR})^T, \quad (4)$$

where $\pi$ is the camera perspective projection function, $T_{CW} \in \mathbf{SE}(3)$ is the camera pose of a viewpoint in the world coordinate system. $\mathbf{J}$ is the Jacobian of the perspective projection function and $\mathbf{R} \in \mathbf{SO}(3)$ is the rotation component of the camera pose $T_{CW}$. Similar to color, a rendered depth for a pixel $p$ can be written as:

$$D(p) = \sum_{i=1}^n d_i^C f(p) \prod_{j=1}^{i-1}(1 - f(p)), \quad (5)$$

where $d_i^C = \left[T_{CW} \mu_i^W\right]_Z$ is the depth, i.e. $Z$ coordinate, of a Gaussian $i$ in camera coordinate frame. This formulation ensures that the rendered Gaussian splats are differentiable with respect to their 3D Gaussian splat parameters. By employing gradient descent optimization, Gaussian splats iteratively refine their optical and geometric parameters, thereby enabling an accurate representation of the scene with high fidelity.

### 3.3. Differentiable camera pose estimation

The formulation of the projected 2D Gaussian splats in (4) ensures that they are differentiable with respect to the camera pose $T_{CW}$ as well. Applying the chain rule to (4):

$$\frac{\partial \mu^I}{\partial T_{CW}} = \frac{\partial \mu^I}{\partial \mu^C} \frac{\partial \mu^C}{\partial T_{CW}},$$

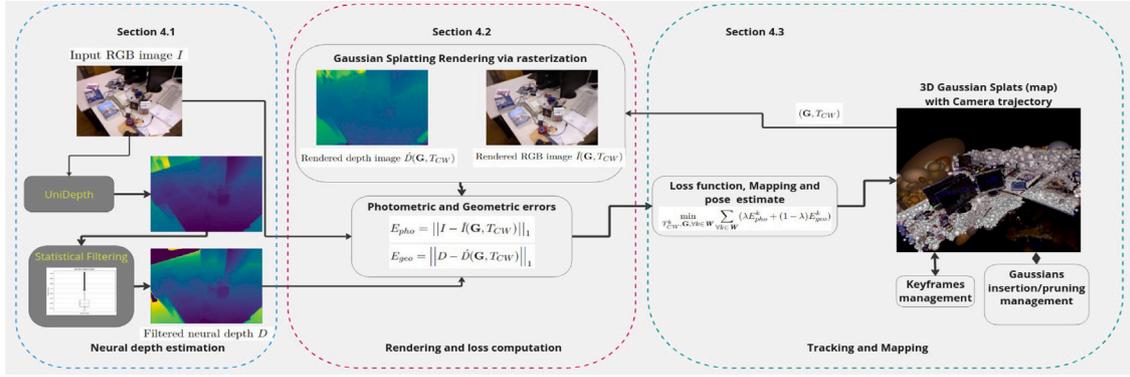

**Fig. 2.** *UDGS-SLAM* pipeline consists of three phases: neural depth estimation and local consistency enforcement (left), image rendering and loss computation (middle), and camera pose estimation and map parameter updates (right).

$$\frac{\partial \Sigma^I}{\partial T_{CW}} = \frac{\partial \Sigma^I}{\partial \mathbf{J}} \frac{\partial \mathbf{J}}{\partial \mu^C} \frac{\partial \mu^C}{\partial T_{CW}} + \frac{\partial \Sigma^I}{\partial \mathbf{R}} \frac{\partial \mathbf{R}}{\partial T_{CW}}, \qquad (6)$$

where $\mu^C$ represents the 3D position of a Gaussian splat in the camera coordinate frame. Following [19], the derivatives with respect to the camera pose $T_{CW}$ are derived using the exponential and logarithmic mapping between Lie algebra and the Lie group as follows,

$$\frac{\partial \mu^C}{\partial T_{CW}} = \mathbf{I} - \Omega^+,$$

$$\frac{\partial \mathbf{R}}{\partial T_{CW}} = \begin{bmatrix} 0 & -\mathbf{R}_1^+ \\ 0 & -\mathbf{R}_2^+ \\ 0 & -\mathbf{R}_3^+ \end{bmatrix}, \qquad (7)$$

where $\Omega^+$ and $\mathbf{R}_i^+$ represent the skew matrices of $\mu^c$ and the $i$th column of $\mathbf{R}$, respectively.

## 4. SLAM pipeline

This section presents the details of the UDGS-SLAM pipeline. An overview of the system is summarized in Fig. 2.

### 4.1. Neural depth estimation

The UniDepth network is utilized to estimate the scene depth from a single RGB image captured by a monocular camera [27]. The network extracts color features from the RGB image and predicts the depth values based on these features. UniDepth employs different backbone encoders, with our findings showing that the ViT-style large-size model encoder delivers the highest accuracy. A performance comparison among different backbones is presented as an ablation study in Section 6. Regardless of the backbone, the estimated depth image is not locally consistent. Similar to stereo-depth estimation [58], the UniDepth estimated values exhibit a left-skewed pattern with heavy right tails, as shown in Fig. 3.c. These heavy tail patterns are particularly noticeable at transitions between proximal and distal objects (see Fig. 3.a and b). In the Gaussian Splatting framework, the color features (from the RGB image) and the predicted depth image from the Unidepth network are merged by associating each pixel's color with its corresponding depth estimate. This combined input is then used to optimize the parameters of the 3D Gaussians, ensuring consistency in both spatial geometry (depth) and appearance (color). Empirical observations have indicated that filtering out these extreme values and ensuring local consistency in the depth map can enhance trajectory and map estimation accuracy. To assure local consistency, We introduce a straightforward yet effective statistical filtering method. The method retains only those depth values falling within the Interquartile Range (IQR), and marks any outliers beyond this range as invalid. Subsequently, only the valid depth values are utilized for geometric error computation. The use of (IQR) for filtering depth estimates is justified by its robustness to outliers without assuming data normality, making it particularly suitable for depth data with non-Gaussian noise distributions. IQR-based filtering offers computational efficiency critical for real-time SLAM systems while providing comparable performance to more complex methods [59]. Due to its practicality, IQR has been effectively applied for outliers rejection in similar application [60,61], where its balance of robustness and efficiency outweighs the theoretical advantages of computationally intensive alternatives like deep learning-based outlier rejection. After applying statistical filtering to the depth image, a local consistent depth image is obtained, with outliers marked as invalid values, as presented in Fig. 3.d. The ablation studies highlight the importance of the statistical filter, as the performance degrades when it is not used, as shown in Table 4.

### 4.2. Rendering and loss computation

The 3D Gaussians $\mathbf{G}$ can be rendered for a viewpoint of given camera pose $T_{CW}$ via differentiable rasterization. Rasterization involves sorting and alpha-blending of the Gaussians as outlined in Section 3.2. The derivatives for all parameters are calculated explicitly. We have adopted the implementations from [19,62] for rendering RGB and depth images, respectively. Once an RGB image is rendered, the photometric error is calculated by comparing the rendered image to the captured image as follows,

$$E_{pho} = \left\| I - \hat{I}(\mathbf{G}, T_{CW}) \right\|_1, \qquad (8)$$

where $I$ is the input frame and $\hat{I}(\mathbf{G}, T_{CW})$ is the rendered rgb image of the Gaussians $\mathbf{G}$ at a view point of a given camera pose $T_{CW}$. The rendered RGB image can be computed as explained in Eq. (3). Similarly, the geometric error can be computed as

$$E_{geo} = \left\| D - \hat{D}(\mathbf{G}, T_{CW}) \right\|_1, \qquad (9)$$

where $D$ is the filtered neural depth calculated as presented in Section 4.1 and $\hat{D}(\mathbf{G}, T_{CW})$ is the rendered depth image. The rendered depth image can be computed as explained in Eq. (5). A total loss function can then be formulated from a weighted combination of the photometric and geometric errors as follows,

$$\mathcal{L}(\mathbf{G}, T_{CW}) = \lambda E_{pho} + (1 - \lambda) E_{geo}, \qquad (10)$$

where $\lambda$ is a weighting factor that balances the contribution of the photometric error $E_{pho}$ and the geometric error $E_{geo}$ in the total loss.

### 4.3. Tracking and mapping

This section introduces the different steps used for refining 3D Gaussian splats (map) and camera pose.

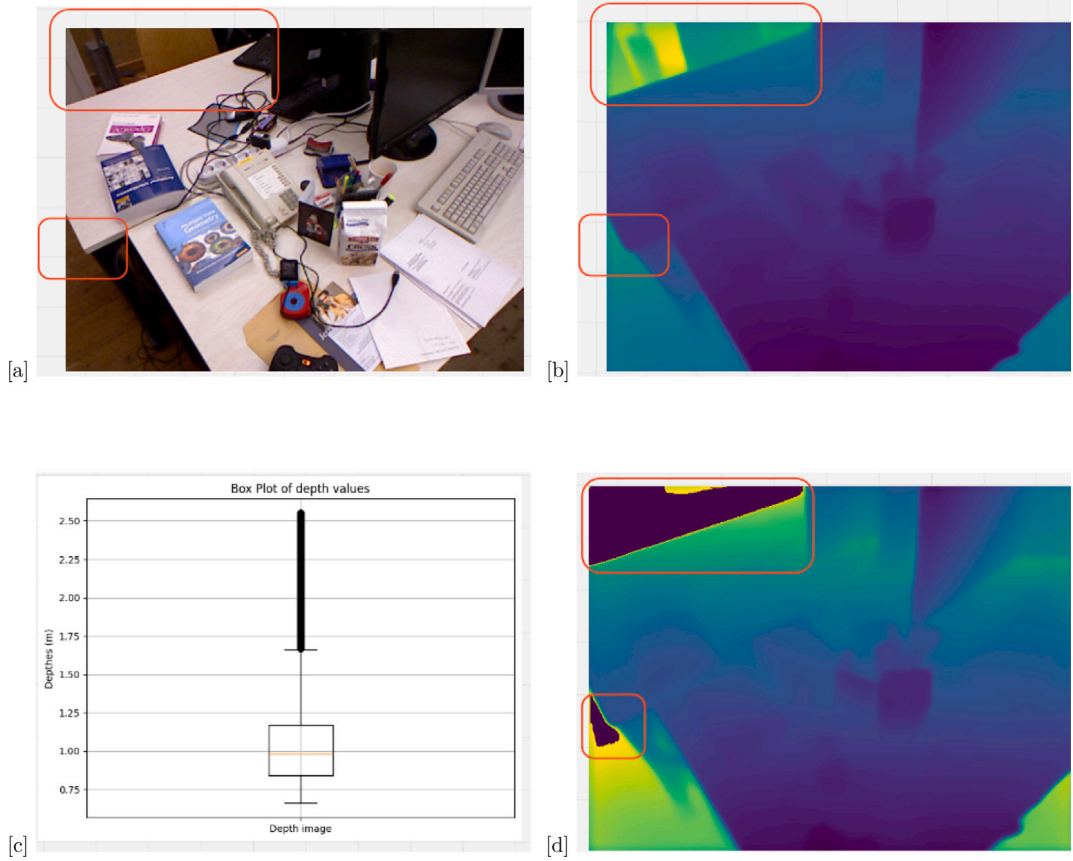

**Fig. 3.** The UniDepth network is used to estimate the scene depth from a single RGB image(a). The depths at transitions between proximal and distal objects exhibit nonconsistency (b). This inconsistency makes a lift-skewed distribution with right heavy tails (c). By applying statistical filtering, local consistency is achieved by marking outliers as invalid values (d).

### 4.3.1. Keyframes management

Although it is theoretically possible to use all previously obtained RGB and depth images for refining the map parameters and camera poses, this method is practically infeasible due to computational constraints. Instead of using all the images, carefully selected keyframes within a small window $\mathcal{W}_k$ are used. The keyframes should not be redundant, should observe the same area [63], and should maintain a wide baseline among them to provide robust multiview constraints [64, 65]. following Matsuki et al. [19] and DSO [63], a small window $\mathcal{W}_k$ of keyframes is maintained. A new frame is considered to be a keyframe by assessing its covisibility, which is calculated by determining the intersection over union of the observed Gaussians between the current frame and the previous keyframe. A new keyframe is added to the window if the covisibility falls below a certain threshold or if the translation (baseline) between the current frame and the previous keyframe is significantly large relative to the median depth.

### 4.3.2. Gaussians insertion and pruning management

As the camera moves, newly unobserved areas come into view, requiring the insertion of new Gaussians to capture their optical and geometric properties. New Gaussians are inserted at each keyframe for these newly observed regions, with their means $\mu^w$ initialized by back-projecting the filtered UniDepth-estimated depth values. Their optical properties are derived from the corresponding RGB input image, and an initial opacity of 0.5 is assigned to ensure stable initialization for gradient-based optimization, preventing any single splat from dominating the rendering process. Instead of inserting one Gaussian per pixel, a structured placement strategy is employed to reduce redundancy while preserving fine details. An adaptive density control mechanism prioritizes insertions in high-gradient areas while minimizing unnecessary additions in low-texture regions, optimizing computational efficiency without sacrificing accuracy. The newly inserted and existing Gaussians are refined through sequential optimization, adjusting their properties dynamically. In addition to Gaussian insertion, excess Gaussians are pruned; if a Gaussian within a keyframe is not observed in at least three subsequent frames, it is considered geometrically unstable and removed from the scene.

### 4.3.3. Tracking and mapping

The purpose of the tracking and mapping module is to maintain a 3D Gaussian map of the scene where each Gaussian is defined as explained in Eq. (1) and to estimate the camera pose for each obtained RGB frame. In addition, the map should be coherent and consistent enough to allow rendering RGB images with high fidelity. To achieve this consistency, a window of previously obtained keyframes $\mathcal{W}_r$ is used along with the current window of keyframes $\mathcal{W}_k$ for map and poses refinement. Similar to [19], two past keyframes are selected randomly to form $\mathcal{W}_r$. The 3D Gaussian parameters (map parameters) and the camera pose estimation are formulated as an optimization problem and their parameters can be estimated by minimizing the loss function in (10) as follows,

$$\min_{T_{CW}^k, \mathbf{G}, \forall k \in \mathcal{W}} \sum_{\forall k \in \mathcal{W}} (\lambda E_{pho}^k + (1-\lambda) E_{geo}^k), \qquad (11)$$

where $k$ is a keyframe and $\mathcal{W} = \mathcal{W}_r \cup \mathcal{W}_k$ is an optimization window of keyframes, calculated as the union of the randomly selected previous keyframes $\mathcal{W}_r$ and the keyframes in the current window $\mathcal{W}_k$.

### 4.4. Pipeline initialization

Unlike Matsuki et al. in their monocular camera pipeline [19], UDGS-SLAM does not use any prior information about scene depth in the initialization step. Instead, the Gaussians are initialized at the depths of the filtered UniDepth estimated depth image. Their color proprieties are obtained from the corresponding pixels in the input RGB image. Then, The Gaussians parameters are refined further by minimizing the loss function in (10) using gradient descent for the map parameters solely. During initialization, the initial camera pose $T_{CW}$ is set to $[\mathbf{I}_{3x3}|\mathbf{0}_{3x1}]$ or to the pose of the camera in the world coordinate system if it is known.

## 5. Experiments and results

An evaluation of the proposed system is conducted on real-world dataset. Additionally, an ablation study is proposed to justify the design choices and to investigate the impact of different UniDepth backbone encoders on the results. This section presents the experiment setup and the results while the ablation study is presented in the subsequent section (Section 6).

### 5.1. Experiment setup

#### 5.1.1. Dataset
The proposed approach is evaluated on TUM RGB-D dataset [66] (3 sequences). Although the dataset includes depth images, RGB images are only used in the proposed approach. Camera pose estimates are compared with the provided ground truth poses. For the ablation study, only one sequence (fr1-desk) from the dataset is used to assess the performance variations among different backbone encoders.

#### 5.1.2. Implementation details
UDGS-SLAM is tested on a laptop with Intel Core i7-13700H, 5.0 GHz, 32 GB RAM, and a single Nvidia GeForce RTX 4070 GPU. 3D Gaussian rendering relies on CUDA C++ implementation proposed at [14,19]. The rest of the pipeline is developed with Pytorch. UDGS-SLAM was able to achieve 5 FPS during the experiments.

#### 5.1.3. Metrics
For camera pose estimation, the pipeline reports the Root Mean Square Error of Absolute Trajectory Error (ATE RMSE ↓) of the estimated keyframes. To evaluate map and rendering quality, the pipeline reports standard metrics: Peak Signal-to-Noise Ration (PSNR ↑), Structural Similarity Index Measurement (SSIM ↑), and Learned Perceptual Image Patch Similarity (LPIPS ↓) [50].

#### 5.1.4. BaseLine methods
Since UDGS-SLAM does not incorporate loop closure, it is compared with similar SLAM methods that also lack explicit loop closure routines, which include NeRF and Gaussian splatting-based methods that achieve photorealistic map representation. Given the proposed solution reliance on monocular images, it is benchmarked against other monocular-based Gaussian splatting solutions. Due to the scarcity of monocular-based Gaussian splatting SLAM solutions, RGB-D Gaussian splatting methods are also considered for a more comprehensive performance comparison. Specifically, for RGB based methods the proposed solution is compared with DROID-VO [67]. DepthCov-VO [16], and MonoGS [19] using its monocular implementation. For RGB-D based methods, the proposed solution is compared with NICE-SLAM [17], Vox-Fusion [18], and SplaTAM [20].

### 5.2. Evaluation

This section presents the results of UDGS-SLAM. The results discussed herein utilize the ViT large model encoder of the UniDepth network and a statistical filter to ensure local consistency. The ablation study section discusses other UniDepth encoder backbones with/without statistical filtering (see Section 6).

#### 5.2.1. Camera tracking accuracy
Fig. 4 presents camera trajectory estimation for 3 sequences. Despite poor RGB image quality (resolution is 640 × 480) and high motion blur, UDGS-SLAM gives small ATE RMSE. In Table 1, UDGS-SLAM's camera pose estimation is benchmarked against various baselines using the TUM RGB-D dataset. A comprehensive quantitative analysis reveals that the proposed method performs well against both Gaussian splatting and non-Gaussian splatting-based methods. Additionally, The comparisons also include methods utilizing both monocular and RGB-D inputs. Remarkably, the proposed approach not only outperforms other monocular-based methods but also exceeds the performance of RGB-D-based methods. It reports the best (lowest) ATE RMSE, achieving the minimum trajectory error compared to all baselines in the fr1-desk sequence - reducing the error by more than 10% from 3.5 cm to 3.0 cm compared to SplaTAM [20]. In the fr2-xyz sequence, it achieves the second-best performance among monocular-based methods, trailing only behind DepthCov-VP [16], and outperforms the RGB-D based method NICE-SLAM [17]. In the fr3-office sequence, although the method surpasses some baselines (DepthCov-VO [16] and Vox-Fusion [18]), its performance is not as strong compared to other sequences. This may be attributed to high motion blur caused by fast camera motion and the larger covered area in the fr3-office sequence, which leads to drift in the estimated trajectory. This highlights potential areas for future improvement.

#### 5.2.2. Rendering results
In addition to camera pose tracking estimation, the rendering performance is also analyzed for high photorealistic reconstruction. In UDGS-SLAM, the scene/map is represented by a number of Gaussians as explained in Section 3.1 similar to the depiction in Fig. 1.d. For a given viewpoint of camera pose, the scene can be rendered to produce a photorealistic image similar to the one presented in Fig. 1.f. Using the metrics in Section 5.1.3, Table 2 reports the rendering performance of UDGS-SLAM on TUM dataset showing good rendering metrics across all sequences. The rendering metrics for the fr2-xyz and fr3-office scenarios are higher than those for fr1-desk. This is because the former scenarios feature distinct, non-occluded objects with sufficient separation and varying depths, whereas fr1-desk contains many occluded objects at similar distances.

UDGS-SLAM rendering metrics are compared with several baselines. The average metrics are reported in Table 3. It is worth noting that the rendering metrics of SplaTAM for TUM RGB-D dataset were not reported by its authors [20].

These results clearly demonstrate that the proposed approach not only surpasses monocular-based methods but also outperforms the RGB-D-based Point-SLAM. Furthermore, it achieves comparable performance, closely matching that of MonoGS in its RGB-D configuration, while using only a monocular camera.

## 6. Ablation study

In Table 4, an ablative analysis is conducted to validate the design choices of the UniDepth network. The study examines the performance of different encoder backbones for the UniDepth network, both with (w) and without (w/o) statistical filtering to ensure local consistency. This analysis includes both version 1 (V1) and version 2 (V2) architectures of UniDepth [46,68]. For V1, the ViT Large model and ConvNext are used as backbone encoders. For V2, the ViT Large model, the only available encoder at the time of testing, is utilized. The evaluation is conducted on the fr1-desk sequence of the TUM dataset. The results indicate that the UniDepth V1 network, when combined with the ViT Large model and statistical filtering, achieves the lowest ATE-RMSE and the highest rendering metrics.

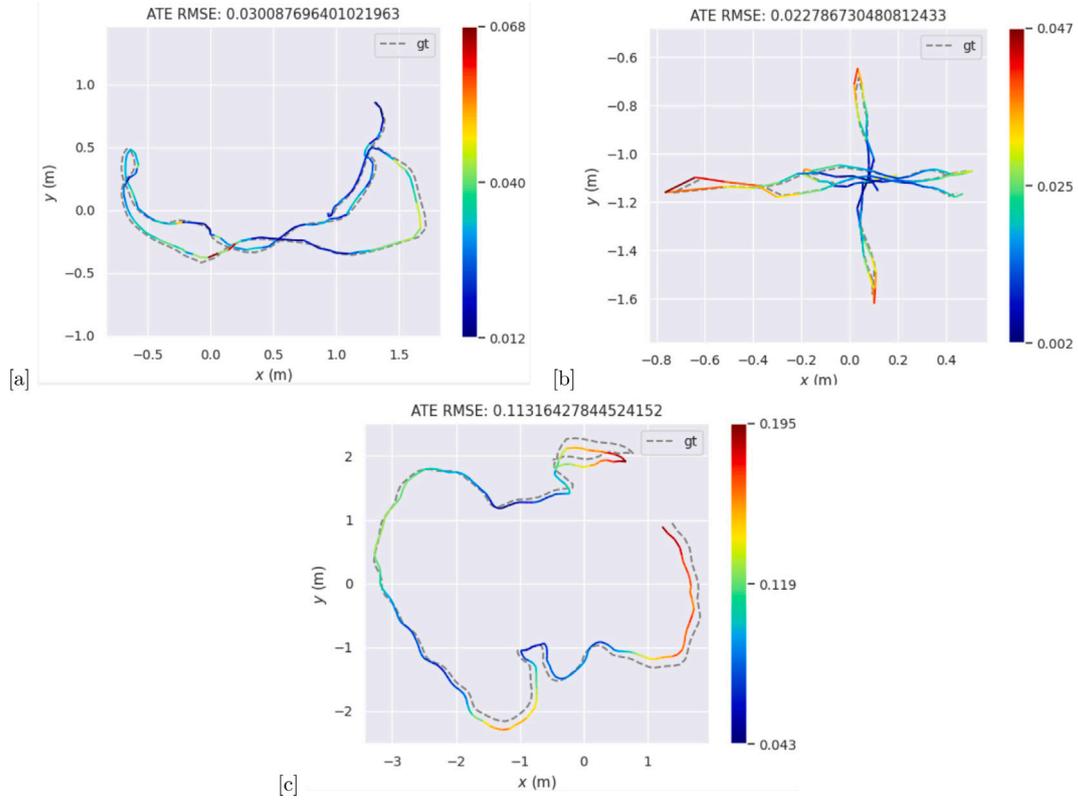

**Fig. 4.** ATE RMSE (↓m) of Camera pose estimation using UDGS-SLAM. (a) Fr1-desk sequence trajectory estimation. (b) Fr2-xyz sequence trajectory estimation. (c) Fr3-office sequence trajectory estimation.

**Table 1**
Camera tracking results on TUM for monocular and RGB-D. ATE RMSE in (↓cm) is reported.

| Methods | Input | Based on | fr1-desk | fr2-xyz | fr3-office |
| --- | --- | --- | --- | --- | --- |
| DROID-VO [67][a] | Monocular | ConvGRU | 5.2 | 10.7 | 7.3 |
| DepthCov-VO [16][a] | Monocular | Gaussian Process | 5.6 | 1.2 | 68.8 |
| MonoGS [19][a] | Monocular | Gaussian Splatting | 3.78 | 4.6 | 3.5 |
| NICE-SLAM [17][a] | RGB-D | NERF | 4.26 | 6.19 | 3.87 |
| Vox-Fusion [18][a] | RGB-D | NERF | 3.52 | 1.49 | 26.01 |
| SplaTAM [20] | RGB-D | Gaussian Splatting | 3.35 | 1.24 | 5.16 |
| **UDGS-SLAM(ours)** | Monocular | Gaussian Splatting | 3.0 | 2.2 | 11.3 |

[a] The results are adapted from [19].

**Table 2**
UDGS-SLAM rendering metrics for TUM dataset.

| Metric | fr1-desk | fr2-xyz | fr3-office | Average |
| --- | --- | --- | --- | --- |
| PSNR ↑ | 23.3 | 24.9 | 23.6 | 24 |
| SSIM ↑ | 0.79 | 0.8 | 0.806 | 0.8 |
| LPIPS ↓ | 0.26 | 0.22 | 0.26 | 0.246 |

**Table 3**
Average rendering metrics for TUM dataset.

| Method | Input | PSNR ↑ | SSIM ↑ | LPIPS ↓ |
| --- | --- | --- | --- | --- |
| MonoGS [19] | Monocular | 21 | 0.7 | 0.3 |
| MonoGS [19][a] | RGB-D | 24.37 | 0.804 | 0.225 |
| Point-SLAM [50][a] | RGB-D | 21.39 | 0.727 | 0.463 |
| SplaTAM [20] | RGB-D | – | – | – |
| **UDGS-SLAM (ours)** | Monocular | 24 | 0.8 | 0.246 |

[a] The results are adapted from [19].

## 7. Conclusion

This work presents UDGS-SLAM, a system that adapts 3D Gaussians as its underlying map representation, enabling photorealistic rendering, dense mapping, and camera trajectory optimization without the need for explicit prior knowledge about the scene or camera motion. UDGS-SLAM leverages advances in neural depth estimation from a single RGB image by utilizing depth maps generated by the UniDepth network. Additionally, it employs a straightforward yet effective statistical filtering method to ensure local consistency and enhance estimation and rendering accuracy. The effectiveness of UDGS-SLAM is demonstrated through testing on the TUM RGB-D dataset, where it exhibits competitive performance, achieving results comparable to or better than existing baselines. This work highlights the potential of integrating neural depth estimation from monocular cameras with Gaussian splatting to develop more sophisticated and efficient SLAM methods. Nonetheless, potential improvements in the proposed approach remain. For example, due to their complementary nature, integrating image-IMU depth estimation with neural depth could yield more accurate depth maps, thereby enhancing overall performance. Furthermore, exploring the incorporation of loop closure could increase the global consistency of the map. These aspects will be investigated in future work. Additionally, dynamic scenes can be handled by incorporating motion segmentation to identify dynamic objects. This can be followed by selective filtering or separate modeling of static and dynamic components, allowing for the exclusion of dynamic Gaussians from the mapping process. Alternatively, dynamic objects can be explicitly modeled using separate sets of Gaussians that incorporate velocity parameters.

**Table 4**
Ablation analysis on the UniDepth backbone encoders with/without local consistency. The analysis confirms that using a V1-ViT large model encoder with statistical filtering to assure local consistency gives the best performance.

| Backbone encoder | Statistical filtering | ATE-RMSE ↓cm | PSNR ↑ | SSIM ↑ | LPIPS ↓ |
|---|---|---|---|---|---|
| V1-ViT large | w | 3 | 23.3 | 0.79 | 0.26 |
|  | w/o | 3.6 | 23 | 0.75 | 0.27 |
| V1-ConvNext large | w | 3.7 | 21 | 0.7 | 0.28 |
| V2-ViT large | w | 6.2 | 23 | 0.75 | 0.28 |
|  | w/o | 5.8 | 23 | 0.7 | 0.25 |

**CRediT authorship contribution statement**

**Mostafa Mansour:** Writing – original draft, Visualization, Validation, Software, Resources, Methodology, Investigation, Formal analysis, Data curation, Conceptualization. **Ahmed Abdelsalam:** Writing – review & editing, Writing – original draft, Visualization, Validation, Software, Methodology, Investigation, Formal analysis, Data curation. **Ari Happonen:** Supervision, Project administration, Funding acquisition. **Jari Porras:** Supervision, Project administration, Funding acquisition. **Esa Rahtu:** Writing – review & editing, Supervision, Project administration.

**Declaration of competing interest**

The authors declare that they have no known competing financial interests or personal relationships that could have appeared to influence the work reported in this paper.

**Data availability**

Data will be made available on request.